\documentclass[10pt,twocolumn,letterpaper]{article}

\usepackage[pagenumbers]{cvpr} % To force page numbers, e.g. for an arXiv version

\usepackage{graphicx}
\usepackage{times}
\usepackage{helvet}
\usepackage{courier}
\usepackage{amsmath}
\usepackage{algorithm}
\usepackage{algorithmic}
\usepackage{csquotes} 
\usepackage{color}
\usepackage{paralist}
\usepackage{amssymb}
\usepackage{indentfirst}
\usepackage{pifont}
\usepackage{float}
\usepackage{multirow}
\usepackage{cite}
\usepackage{mathrsfs}
\usepackage{makecell}

\usepackage{mathrsfs}
\usepackage{textcomp,booktabs}
\usepackage{amssymb}% http://ctan.org/pkg/amssymb
\usepackage{pifont}% http://ctan.org/pkg/pifont
\usepackage[misc]{ifsym}
\usepackage{ulem}

\usepackage{soul}
\usepackage{colortbl}
\usepackage[dvipsnames]{xcolor}
\usepackage{mathrsfs}
\usepackage{color}
\usepackage{xcolor}
\definecolor{citecolor}{HTML}{0071bc} 
\definecolor{SeaGreen4}{RGB}{0,205,102} 
\definecolor{SlateBlue}{RGB}{106,90,205} 
\definecolor{DarkRed}{RGB}{178,34,34} 
\usepackage[colorlinks, linkcolor=red,  anchorcolor=blue, citecolor=citecolor]{hyperref}

\usepackage{colortbl}
\definecolor{mygray}{gray}{.9}
\definecolor{mypink}{rgb}{.99,.91,.95}
\definecolor{mycyan}{cmyk}{.3,0,0,0}

\usepackage{color}
\usepackage{xcolor}
\definecolor{citecolor}{HTML}{0071bc} 
\definecolor{SeaGreen4}{RGB}{0,205,102} 
\definecolor{SlateBlue}{RGB}{106,90,205} 
\definecolor{DarkRed}{RGB}{178,34,34}

\usepackage[capitalize]{cleveref}
\crefname{section}{Sec.}{Secs.}
\Crefname{section}{Section}{Sections}
\Crefname{table}{Table}{Tables}
\crefname{table}{Tab.}{Tabs.}

\def\paperID{9748}      % *** Enter the Paper ID here
\def\confName{CVPR}
\def\confYear{2024}

\title{ CXPMRG-Bench: Pre-training and Benchmarking for X-ray Medical Report Generation on CheXpert Plus Dataset } 

\author{Xiao Wang$^{1}$, Fuling Wang$^{1}$, Yuehang Li$^{1}$, Qingchuan Ma$^{1}$, Shiao Wang$^{1}$, \\  Bo Jiang$^{1}$\thanks{\Letter~~Corresponding Author: Bo Jiang}, Chuanfu Li$^{2}$, Jin Tang$^{1}$ \\ 
${^1}$ {School of Computer Science and Technology, Anhui University, Hefei, China} \\
${^2}$ {First Affiliated Hospital of Anhui University of Chinese Medicine, Hefei, China} \\ 
\textit{\{xiaowang, jiangbo, tangjin\}@ahu.edu.cn, licf@ahtcm.edu.cn} \\ 
\textit{\{e23201049, e23201112, e02114334\}@stu.ahu.edu.cn, wsa1943230570@126.com} 
}

\begin{document}
\maketitle

\begin{figure*}[!htp]
\centering
\includegraphics[width=\linewidth]{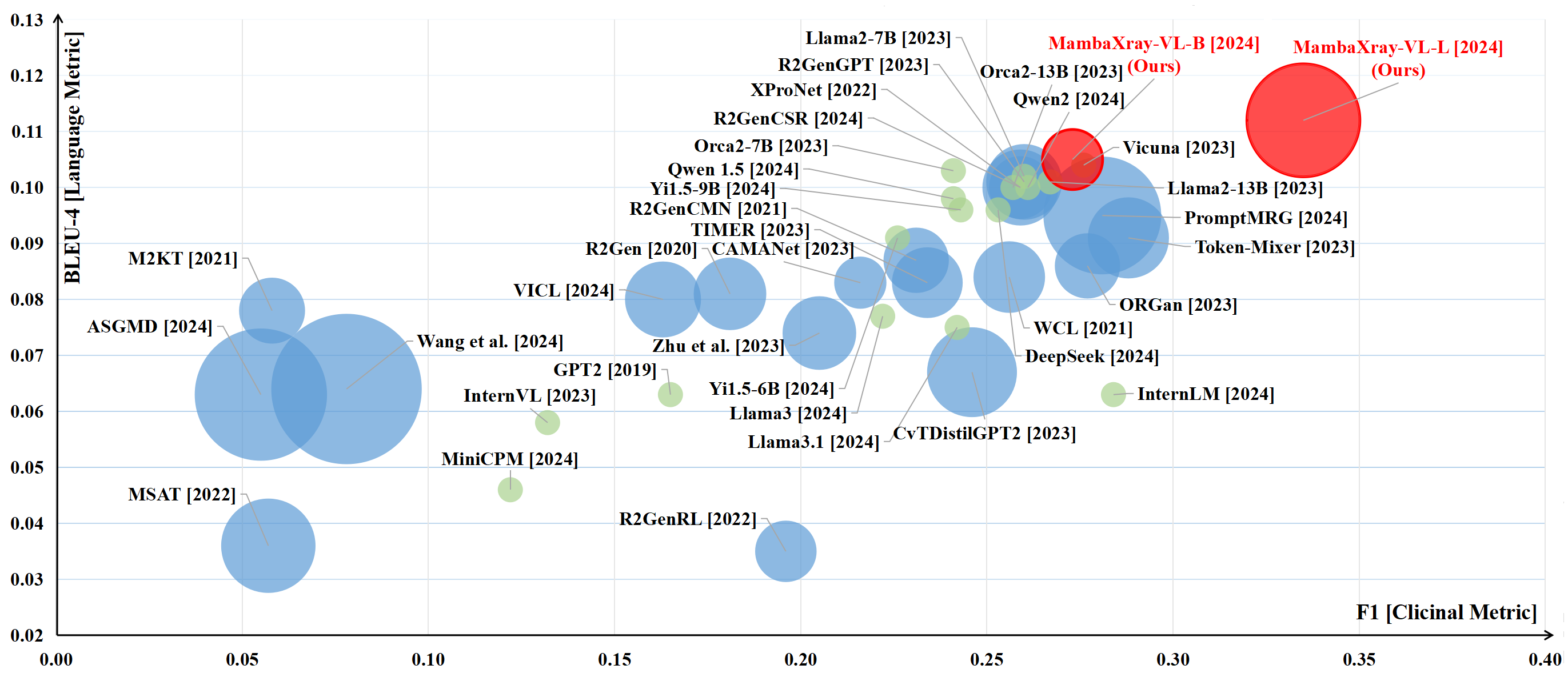}
\caption{
An overview of the benchmarked LLM/VLM-based (\textcolor[RGB]{169,209,142}{green circle}) and mainstream MRG models (\textcolor[RGB]{91,155,213}{blue circle}) on the CheXpert Plus dataset in this paper. 
}
\label{fig:firstIMG}
\end{figure*}

%%%%%%%%% ABSTRACT
\begin{abstract}
X-ray image-based medical report generation (MRG) is a pivotal area in artificial intelligence which can significantly reduce diagnostic burdens and patient wait times. Despite significant progress, we believe that the task has reached a bottleneck due to the limited benchmark datasets and the existing large models' insufficient capability enhancements in this specialized domain. Specifically, the recently released CheXpert Plus dataset lacks comparative evaluation algorithms and their results, providing only the dataset itself. This situation makes the training, evaluation, and comparison of subsequent algorithms challenging. Thus, we conduct a comprehensive benchmarking of existing mainstream X-ray report generation models and large language models (LLMs), on the CheXpert Plus dataset. We believe that the proposed benchmark can provide a solid comparative basis for subsequent algorithms and serve as a guide for researchers to quickly grasp the state-of-the-art models in this field. More importantly, we propose a large model for the X-ray image report generation using a multi-stage pre-training strategy, including self-supervised autoregressive generation and Xray-report contrastive learning, and supervised fine-tuning. Extensive experimental results indicate that the autoregressive pre-training based on Mamba effectively encodes X-ray images, and the image-text contrastive pre-training further aligns the feature spaces, achieving better experimental results. 
%% 
% All the source codes will be released upon acceptance. 
%% 
Source code can be found on \url{https://github.com/Event-AHU/Medical_Image_Analysis}.  
\end{abstract}

% GMAI-MMBench: A Comprehensive Multimodal Evaluation Benchmark Towards General Medical AI 

\section{Introduction}

X-ray image based Medical Report Generation (MRG) is a critical research problem in artificial intelligence, which targets describing the \textit{findings} or \textit{impressions} from the given X-ray data using natural language. The successful implementation of this task can significantly reduce the diagnostic burden on physicians, decrease patient wait times, and foster the positive application of artificial intelligence. However, the path to progress in this direction is not smooth sailing, there remain formidable challenges that need to be overcome. The challenging issues include image interpretation, data annotation, heterogeneity issues, consistency and standardization of reports, diversity and variability of diseases, interpretability of algorithms, etc. How to address these challenges further and improve the quality of medical report generation remains an urgent research problem.

After revisiting the mainstream algorithms of X-ray image medical report generation, we find that datasets like IU X-ray and MIMIC-CXR are widely used for the training and evaluation of report generation models. However, the IU X-ray only contains 7,470 images and 3,955 radiology reports samples, which is rather limited, especially in the large model era. The recently released CheXpert Plus dataset~\cite{chambon2024CheXpertPLUS} is a large-scale dataset for the X-ray report generation, however, they did not release comparative methods, making it difficult for subsequent algorithms to conduct experiments and comparisons on this dataset. Therefore, we conduct a comprehensive benchmarking of existing open-sourced mainstream X-ray report generation models, Large Language Models (LLMs), and Vision-Language Models (VLMs), termed \textbf{CXPMRG-Bench}, on the newly released CheXpert Plus dataset, as shown in Fig.~\ref{fig:firstIMG}. The completion of this work can also help researchers identify which large models and algorithms are currently leading in the field of X-ray report generation.

On the other hand, most mainstream algorithms follow the encoder-decoder framework which usually adopts the vision encoder (e.g., ResNet~\cite{He2016resnet}, Transformer~\cite{Ashish2017Transformer}) to process the given X-ray data and a text decoder (e.g., LSTM~\cite{Sepp1997LSTM}, GRU~\cite{Chung2014GRU}, Transformer~\cite{Ashish2017Transformer}) for report generation. Along with the development of pre-trained LLM and VLM, the quality of medical reports is enhanced significantly. There are already some researchers who exploit the pre-training for the X-ray report generation. For example, Wang et al.~\cite{wang2024pretraininghighdefinitionxray} propose high-definition X-ray vision models using context-aware masked auto-encoder.  CXR-CLIP~\cite{You2023CXR-Clip} is a new pre-training method that generates more image-text pairs and introduces contrastive loss to enhance the discriminative power of images and texts, effectively learning features in the CXR domain. PTUnifier~\cite{Chen2023PTUnifier} proposes a simple and effective method that utilizes visual and textual prompt pools to make the model compatible with different types of inputs, thereby unifying the advantages of fusion encoders and dual encoders.  
However, we believe these models may be limited by the following issues: 
\textit{Firstly}, the Transformer vision backbone brings huge computational costs $\mathcal{O}(N^2)$, which is not hardware friendly; 
\textit{Secondly}, many X-ray models are pre-trained in a single stage, which may constrain their overall performance. As pure X-ray images are abundant and readily collectible, paired X-ray and report data are relatively scarce. Failing to utilize these visual data resources would be a significant missed opportunity.

To address the issues mentioned above, in this work, we exploit multi-stage pre-training for the X-ray image MRG task and propose the \textbf{MambaXray-VL} large model, including \textit{self-supervised autoregressive generation} and \textit{Xray-report contrastive learning}, and \textit{supervised fine-tuning} on each downstream report generation datasets, as shown in Fig.~\ref{fig:framework}. Specifically speaking, we first partition and feed the X-ray image into the Mamba network to predict the next tokens based on previous context tokens in an autoregressive generation manner. This will enhance the vision perception ability of X-ray significantly using the relatively low-cost Mamba network ($\mathcal{O}(N)$). For the second stage, we feed the paired X-ray image and corresponding reports into the Mamba vision backbone and text encoder (Bio\_ClinicalBERT~\cite{Emily2019Bio_ClinicalBERT}, Llama2~\cite{touvron2023llama2}) for contrastive learning. It will align the X-ray image and reports using the pre-trained feature space. After that, we conduct supervised fine-tuning on each downstream X-ray report generation dataset to achieve higher performance by feeding the X-ray image into the pre-trained Mamba vision backbone network and LLM decoder network. Extensive experiments on three MRG benchmark datasets demonstrate that our pre-trained MambaXray-VL model achieves state-of-the-art performance.

To sum up, the contributions of this paper can be summarized as the following three aspects: 

1). We conduct a comprehensive benchmark for the newly released CheXpert Plus dataset~\cite{chambon2024CheXpertPLUS}, termed \textbf{CXPMRG-Bench}, which covers 19 mainstream X-ray medical report generation algorithms, 14 large language models, and 2 vision-language models. To the best of our knowledge, this benchmark is the first large-scale evaluation of the CheXpert Plus dataset, providing subsequent researchers in the field of X-ray report generation with important reference and comparison criteria. 

2). We propose a new pre-trained large model, termed \textbf{MambaXray-VL}, which adopts the Mamba as the vision encoder and the large language model as the text decoder. Unlike conventional complex Transformer vision models, our Mamba architecture, which employs a multi-stage pre-training strategy, has also achieved state-of-the-art performance. 

3). We extend our research to a broader scope by conducting experiments on the IU X-ray and MIMIC-CXR datasets. We perform analytical experiments and visualizations to deepen the understanding of our MambaXray-VL model's performance and its capabilities in generating X-ray medical reports, thereby enhancing the robustness and generalizability of our findings across different datasets.

\textit{The rest of this paper is organized as follows:} 
In section~\ref{sec::relatedworks}, we review the related works to this paper including X-ray medical report generation, pre-trained large models, and state space model. 
We introduce the pre-trained MambaXray-VL large model for the X-ray medical report generation in section~\ref{sec::mambaXrayVL}. 
After that, we introduce the CXPMRG-Bench benchmark on the CheXpert Plus dataset in section~\ref{sec::benchmark}. 
The experimental configurations and analysis are described in section~\ref{sec::experiments}. 
Finally, we conclude this paper and propose possible research directions in section~\ref{sec::conclusion}.

\section{Related Work} \label{sec::relatedworks}

In this section, we will review the related works on X-ray Medical Report Generation, Pre-trained Large Models, and State Space Models. More works can be found in the following surveys~\cite{wang2024SSMsurvey, wang2023MMPTMs, hartsock2024VLMMRGsurvey}.

\subsection{X-ray Medical Report Generation} 
In recent years, X-ray medical report generation has garnered increasing attention. To enhance model performance, researchers have pursued various improvements in different directions. 
Specifically, DCL~\cite{Li2023DCL} introduces a Dynamic Graph at the visual features of medical images, leveraging knowledge to strengthen the feature representation of these images. RGRG~\cite{Tanida2023RGRG} takes a novel approach by using object detection methods to extract lesion regions and then generating text based on these extracted regions, ultimately combining all the text to form the final report. HERGen~\cite{wang2024HERGen} discovers the historical information between medical reports, treating all reports of a patient as a temporally ordered whole. This approach effectively integrates the temporal and causal information of the reports. 
% CoFE~\cite{li2024CoFE} introduces a counterfactual mechanism by dividing medical images into several patches. Each time, it selects one patch and swaps it with the corresponding patch from a negative sample until the model's predicted disease label changes. This identifies the patch as a key pathological region. Then, a learnable prompt is used to guide the large language model to focus on this area, ultimately achieving good performance. 
R2GenGPT~\cite{Wang2023R2GenGPT} replaces the decoder part of the traditional medical report generation framework with a more powerful large language model, achieving improved performance. 
R2GenCSR~\cite{wang2024r2gencsr} is a recently proposed LLM-based framework for X-ray MRG which employs the Mamba as the visual backbone and retrieves contextual samples from the training set to enhance feature representation and discriminative learning.

It is evident that the vision encoders used in these models are all conventional networks pre-trained on ImageNet~\cite{ImageNet}: DCL~\cite{Li2023DCL} employs ViT~\cite{dosovitskiy2020vit}, RGRG~\cite{Tanida2023RGRG} uses ResNet50~\cite{He2016resnet}, HERGen~\cite{wang2024HERGen} utilizes CvT~\cite{Nicolson2023CvT2DistilGPT2}, R2GenGPT~\cite{Wang2023R2GenGPT} incorporates SwinTransformer~\cite{Liu2021swintransformer}, and R2GenCSR~\cite{wang2024r2gencsr} leverages VMamba~\cite{liu2024vmamba}. 
These encoders, pre-trained on non-medical X-ray images, exhibit certain limitations when extracting features from medical X-ray images. In contrast, our proposed MambaXray-VL is pre-trained on millions of datasets and has a natural advantage in the extraction of features from medical images, especially in the task of medical report generation.

\subsection{Pre-trained Large Models}  
% The goal of self-supervised visual representation learning is to learn strong and transferable representations from unlabeled data, including methods such as contrastive learning, position prediction, and masked image modeling.  
The pre-trained language models, vision models, and vision-language models are widely exploited in nowadays. Currently, the widely used MAE~\cite{He2022MAE} (Masked Autoencoders) is a self-supervised learning method for computer vision, known for its scalability and simplicity. 
% In the MAE approach, parts of the input image patches are randomly masked, and the task is to reconstruct these missing pixels. Autoregressive models have traditionally been used in language or speech modeling, and their application in the image domain has been rare. 
Recently, Apple's team proposed AIM~\cite{elnouby2024AIM}, a series of vision models using autoregressive objectives for pretraining, inspired by large language models, demonstrating similar scaling properties. 
% AIM found that the performance of visual features improves with the increase in model capacity and data volume, and the value of the objective function correlates with the model's performance on downstream tasks. 
ARM~\cite{ren2024arm} is a new self-supervised visual representation learning method based on AIM~\cite{elnouby2024AIM} and Mamba~\cite{gu2024mamba}. Through the autoregressive generation based pre-training, the visual capabilities of the Mamba model can be significantly enhanced, outperforming other training strategies in terms of both efficiency and performance. 
% This improvement is due to the effective utilization of the Mamba model's unidirectional recurrent structure, achieving faster training speeds and excellent accuracy, while also successfully scaling to larger model sizes. However, for medical report generation tasks, focusing solely on medical images is inadequate. Existing datasets are in the form of image-report pairs, making it crucial to fully utilize the textual data. 
CLIP~\cite{Alec2021CLIP} (Contrastive Language-Image Pre-Training) jointly trains image and text encoders using contrastive learning. The key idea is to enable the model to understand and process multi-modal data (images and text) through joint training. 
Inspired by these works, our newly proposed MambaXray-VL utilizes autoregressive generation based pre-training, and CLIP pre-training can achieve better results on medical report generation.

\subsection{State Space Model} 
Since its introduction in 2017, Transformer~\cite{Ashish2017Transformer} has quickly become the preferred model framework for researchers due to its strong performance. However, as the model scales and sequences become longer, its limitations have surfaced. One major drawback is the quadratic growth in computational complexity of the self-attention mechanism with increased context length. Mamba~\cite{gu2024mamba} addresses these issues by using Selective State Space Models (SSMs) to improve traditional state space models and incorporating a hardware-aware parallel algorithm for recurrent operations. 
Vim~\cite{zhu2024vim} (Vision Mamba) is the first SSM model adapted for vision tasks. It uses positional embeddings and bidirectional state space models to achieve high performance, particularly on high-resolution images. VMamba~\cite{liu2024vmamba} extends Mamba by providing a global receptive field with linear complexity. MambaMLP~\cite{ren2024arm} is a new architectural component based on Mamba, designed to enhance feature mixing and representation learning by combining Mamba with an MLP, thereby improving performance on visual tasks. The new SSD (State Space Duality) algorithm proposed by Mamba-2~\cite{Dao2024mamba2} can fully utilize matrix multiplication units on modern hardware, making it 2-8 times faster than the vanilla Mamba. 
% Especially for tasks requiring efficient handling of long sequences and large state sizes, Mamba-2 demonstrates significant improvements in efficiency, scalability, and system optimizations, making it a strong competitor to Transformer. 
% The superior performance of Mamba made us think: \textit{Can we replace the popular Transformer-based visual extractor with a Mamba-based visual extractor?} The success of Vim~\cite{zhu2024vim} shows that Mamba is adequate for visual tasks and has a lot of potential.
The successful applications of the Mamba in many computer vision tasks~\cite{wang2024mambaevt, huang2024mambaFETrack, wang2024MambaPAR} inspired us to adapt it to the pre-trained X-ray large model for medical report generation.

\section{MambaXray-VL Large Model}  \label{sec::mambaXrayVL}
In this section, we will first give an overview of our proposed MambaXray-VL large model, then, we will dive into the details of the proposed multi-stage training strategy. Finally, we highlight some implementation details worth noting in the pre-training phase.

\begin{figure*}
    \centering
    \includegraphics[width=0.85\linewidth]{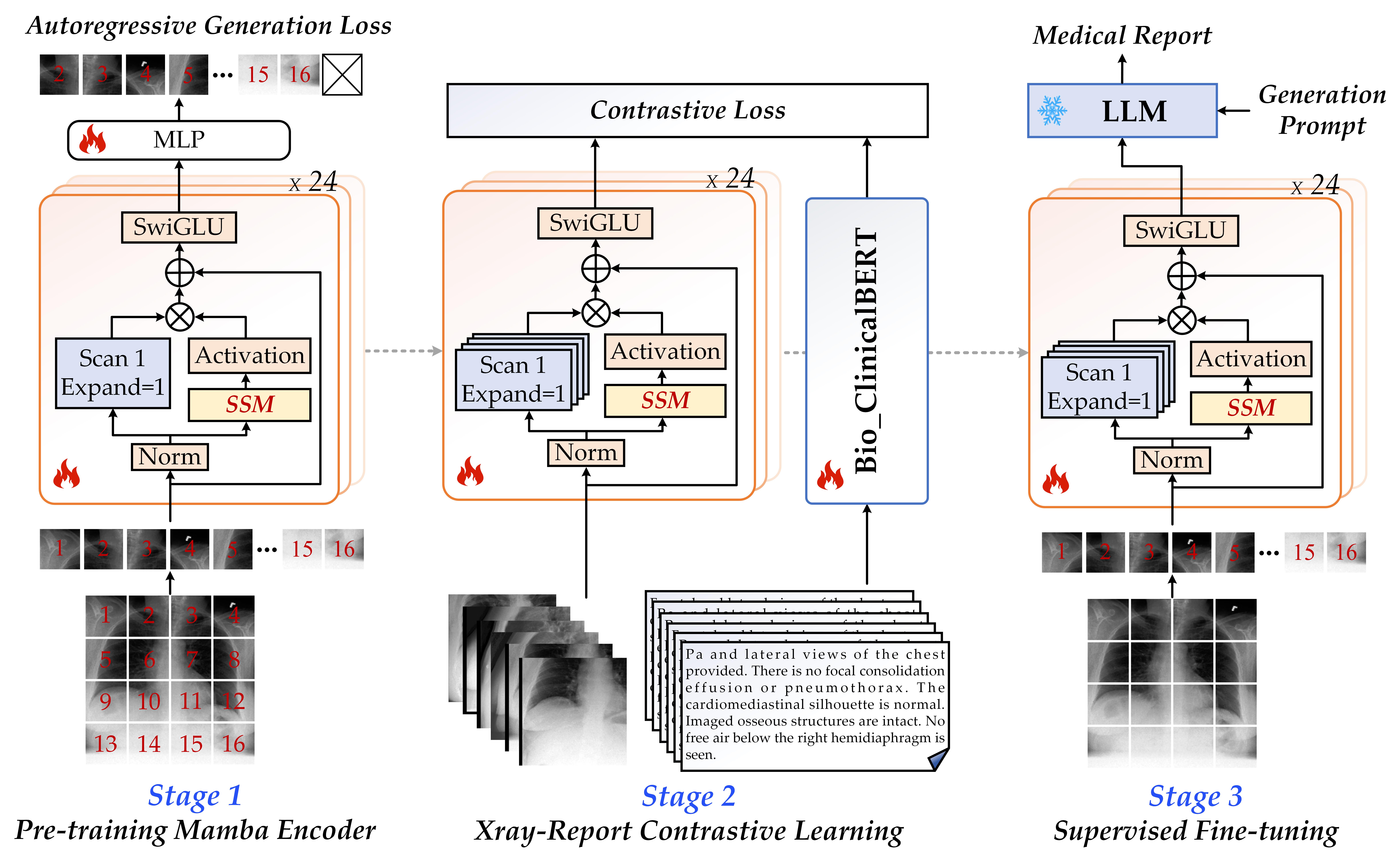}
    \caption{An overview of our proposed MambaXray-VL pre-training framework. It contains three training stages, i.e., Mamba-based autoregressive generation, Xray-report based contrastive learning, and supervised fine-tuning. Specifically, the first phase mainly aims to make full use of larger-scale X-ray visual data to obtain a better visual backbone network (this paper chooses the low-complexity Mamba model). The second phase uses image-text contrastive loss to align X-ray images with medical reports. The third phase can fine-tune on various medical report generation datasets to obtain more refined X-ray report generation results. Note that the layers or modules with \textit{fire}/\textit{snow} symbols denote the parameters that are tuned/frozen in the training phase. 
}
    \label{fig:framework}
\end{figure*}

\subsection{Overview} 
As shown in Fig.~\ref{fig:framework}, we propose a new multi-stage pre-training strategy for the X-ray image medical report generation, including \textit{self-supervised autoregressive generation}, \textit{Xray-report contrastive learning}, and \textit{supervised fine-tuning}. The key insight of multi-stage pre-training instead of joint training is that the aligned Xray-report data are limited, but there are more publicly available X-ray images. Thus, we first pre-training a large-scale vision backbone network on the X-ray images using the Mamba layers, due to a better balance between the computational cost and accuracy. More importantly, we adopt the autoregressive generation to achieve self-supervised learning on the X-ray image. It performs similar or better than the widely used MAE (Masked Auto-Encoder) pre-training strategy for this task. Then, we transfer the Mamba vision backbone to the second stage, i.e., Xray-report contrastive learning. Specifically, we feed the paired data into the pre-trained Mamba vision backbone and language encoder for the vision-language feature extraction. This stage will project the vision and language representations into a shared feature space to bridge the vision-semantic gaps. Finally, we conduct supervised fine-tuning on the training subset of downstream datasets for the X-ray medical report generation.

\subsection{Multi-Stage Pre-training}  

As illustrated in Fig.~\ref{fig:framework}, our proposed \textit{MambaXray-VL} large model contains three training stages which will be introduced in the following paragraphs respectively.

\noindent $\bullet$ \textbf{Stage \#1: Auto-regressive Generation for Mamba Vision Encoder Pre-training.~} 
To make full use of existing X-ray images, we conduct self-supervised learning to obtain a strong vision backbone network. Different from the widely used MAE (Masked Auto-Encoder)-based framework, in this work, we find that the autoregressive generation based framework works similar or even better for the X-ray images, inspired by the success of autoregressive generation in ChatGPT~\cite{openai2023chatGPT}, GPT-4~\cite{openai2024gpt4}, and ARM~\cite{ren2024arm}. 
Let's denote the X-ray image as $\mathcal{I} \in \mathbb{R}^{192 \times 192 \times 3}$, we first partition it into non-overlapping image patches $\mathcal{P}_i \in \mathbb{R}^{16 \times 16 \times 3}, i = \{1, 2, ..., N\}$ and project them into visual tokens $\mathcal{T}_i \in \mathbb{R}^{1024}, i = \{1, 2, ..., N\}$ using a convolutional layer (kernel size $16 \times 16$). Here, \textit{N} is \textit{144} when the resolution of the input X-ray image is set as $3 \times 192 \times 192$. Then, we feed the visual tokens into the \textit{Vim}~\cite{zhu2024vim} backbone network for feature extraction whose complicity $\mathcal{O}(N)$ is much lower than the widely used Transformer $\mathcal{O}(N^2)$. The key operation of \textit{Vim} is the Mamba block (a specific variation of State Space Model~\cite{wang2024SSMsurvey}), as shown in Fig.~\ref{fig:framework}. The visual tokens will first be normalized and fed into the SSM and scan branches. The outputs will be multiplied and added with residual connections. The SwiGLU~\cite{Shazeer2020SwiGLU} is adopted to further process output features before being fed into subsequent Mamba blocks. Finally, an MLP layer is adopted for token reconstruction using the auto-regressive generation loss function.

The objective of autoregressive pre-training is to predict the probability of the next token one by one based on the given corpus $\mathcal{T} = \{ \mathcal{T}_1, \mathcal{T}_2, ..., \mathcal{T}_n \}$, which can be written as:  
\begin{align}
    p(\mathcal{T}) = \prod_{i=1}^{n} p(\mathcal{T}_i | \mathcal{T}_1, ..., \mathcal{T}_{i-1}, \Theta). 
\end{align}
We can find that the likelihood of each token $\mathcal{T}_i$ is computed based on the context of all the proceeding tokens $\{\mathcal{T}_1, ..., \mathcal{T}_{i-1}\}$. Thus, the loss function used for stage 1 can be formulated as follows: 
\begin{align}
    \label{autoregressiveLoss} 
    \mathcal{L}_{AR} = \sum_{i=1}^{n-1} |Vim([\mathcal{T}_1, ..., \mathcal{T}_i]) - \mathcal{T}_{i+1}|^2. 
\end{align}

\noindent $\bullet$ \textbf{Stage \#2: Xray-Report Contrastive Learning.~} 
We adopt the Mamba vision backbone network from the first stage and conduct contrastive learning on the paired Xray-report samples. This will further align the dual modalities as validated in the CLIP~\cite{Alec2021CLIP}. In our implementation, we randomly sample a mini-batch and feed the X-ray images and medical reports into the Vim backbone and the language model (Bio\_ClinicalBERT~\cite{Emily2019Bio_ClinicalBERT}, Llama2~\cite{touvron2023llama2}) and compute the cosine similarity between the paired and unpaired samples: 
\begin{align} 
\label{CTL_loss} 
\mathcal{L}_{CTL} = Similarity(Vim(\mathcal{I}_i), LM(\mathcal{R}_j)), 
\end{align}
where $i$ and $j$ are the index of the X-ray image and report annotation.

\noindent $\bullet$ \textbf{Stage \#3: Supervised Fine-tuning.~} 
After the two pre-training stages, we conduct supervised fine-tuning on the training subset of X-ray image medical report generation. Similar to the first stage, we partition the given X-ray image into non-overlapping patches and project them into visual tokens. Then, the pre-trained Vim backbone network is used for the feature extraction. We concatenate the visual tokens and generation prompt as the input of a large language model for high-performance medical report generation.

In this stage, we adopt the \textit{negative log-likelihood} as the loss function, i.e., 
\begin{align}
    \mathcal{L}_{NLL} = -\sum_{i=1}^{T} log p_{\theta} (y_i|Prompt, [y_1, ..., y_{i-1}]), 
\end{align}
where $\theta$ denotes the trainable parameters and $T$ is the number of words the large language model predicted. 
$Prompt$ is the instruction prompt which is ``\textit{Generate a comprehensive and detailed diagnosis report for this chest X-ray image.}" used in our experiments.

\subsection{Implementation Details}

\noindent $\bullet$ \textbf{Pre-training Stage.} 
Both MambaXray-VL-Base and MambaXray-VL-Large were pre-trained for 100 epochs, with batch sizes set at 256 and 128, respectively. The base learning rate, based on a batch size of 256, was set to 1.5e-4. We adopted a cosine decay schedule with a warm-up for 5 epochs and used the AdamW~\cite{loshchilov2019AdamW} optimizer with a weight decay of 0.05. The resolution of input images is resized to 192 $\times$ 192 in the pre-training phase.

In the second stage, we utilized a vision-text contrastive learning pre-training method to train MambaXray-VL, enabling alignment to the text feature space. Specifically, we used a dataset of 480,000 image-text pairs, composed of publicly available datasets from MIMIC-CXR~\cite{johnson2019mimicCXR}, CheXpert Plus~\cite{chambon2024CheXpertPLUS}, and IU-Xray~\cite{demner2016iuxray}. Inspired by ARM~\cite{ren2024arm}, we used a unidirectional scanning approach in the first stage that fits the autoregressive generation to achieve more efficient pre-training. In the second stage, we extend the scanning block to four copies in order to improve the performance of the model. During this stage, we chose to pre-training for 50 epochs, with a batch size set to 192. The visual encoder was Vim~\cite{zhu2024vim}, loaded with weights from the first stage of pre-training, while the text encoder was Bio\_ClinicalBERT~\cite{Emily2019Bio_ClinicalBERT}, both encoders were set to be trainable. We employed the same optimizer as in the first stage, but the input image size was changed to 224 $\times$ 224.

\begin{table*}
\centering
\caption{Experimental Results on the CheXpert Plus Dataset using \textbf{Mainstream Medical Report Generation Algorithms}. 
\textbf{B4}, \textbf{R}, \textbf{M}, and \textbf{C} is short for BlEU-4, ROUGE-L, METEOR, CIDEr, respectively. \textbf{P}, \textbf{R}, and \textbf{F1} is short for Precision, Recall, F1 score, respectively. \textit{min} is short for minutes. The Param listed in this table denotes the parameters needed to be tuned in the training phase. The best result is highlighted in bold, and the second-best result is underlined. }  
\small 
\resizebox{\textwidth}{!}{ 
\begin{tabular}{c|l|l|l|l|l|l|c|c|c} 
\hline \toprule [0.5 pt] 
\textbf{Index} &\textbf{Algorithm} &\textbf{Publish}  &\textbf{Encoder} &\textbf{Decoder}  &\textbf{\hspace{1ex}B4,\hspace{1em}\hspace{1ex} R,\hspace{1em}\hspace{1ex} M, \hspace{1em} C} &\textbf{\hspace{2ex}P,\hspace{1em}\hspace{1ex} R,\hspace{1em}\hspace{1ex} F1} &\textbf{Time (\textit{min})} &\textbf{Param (\textit{M})} &\textbf{Code} \\
\hline 
\#01  & R2GenRL~\cite{qin2022R2GenRL} & ACL22 & Transformer & Transformer & 0.035, 0.186, 0.101, 0.012 & 0.193, 0.229, 0.196 & 44.33 & 59.87 &   \hyperlink{https://github.com/synlp/R2GenRL}{URL} \\
\hline 
\#02  & XProNet~\cite{wang2022XProNet} & ECCV22 & Transformer &Transformer &  0.100, 0.265, 0.146, 0.121  & 0.314, 0.247, 0.259 & 6.3 & 62.35 &  \hyperlink{https://github.com/Markin-Wang/XProNet}{URL}\\
\hline 
\#03  & MSAT~\cite{wang2022MSAT} & MICCAI22 & ViT-B/16 & Transformer & 0.036, 0.156, 0.066, 0.018 & 0.044, 0.142, 0.057 & 5.72 & 141.10 &  \hyperlink{https://github.com/wang-zhanyu/MSAT}{URL}\\
\hline 
\#04  & ORGan~\cite{hou2023ORGan} & ACL23 & CNN & Transformer  & 0.086, 0.261, 0.135, 0.107 & 0.288, 0.287, 0.277 & 46.66 & 67.50 &  \hyperlink{https://github.com/wjhou/ORGan}{URL}\\
\hline 
\#05  & M2KT~\cite{Yang2023M2KT} &MIA21 & CNN & Transformer & 0.078, 0.247, 0.101, 0.077 & 0.044, 0.142, 0.058 & 22.5 & 69.07 &  \hyperlink{https://github.com/LX-doctorAI1/M2KT}{URL}\\
\hline 
\#06  & TIMER~\cite{Wu2023TIMER} & CHIL23 & Transformer & Transformer & 0.083, 0.254, 0.121, 0.104 & \underline{0.345}, 0.238, 0.234 & 26.5 & 79.28 &  \hyperlink{https://github.com/woqingdoua/TIMER}{URL}\\
\hline 
\#07  & CvT2DistilGPT2~\cite{Nicolson2023CvT2DistilGPT2} &AIM23 & Transformer & GPT2 & 0.067, 0.238, 0.118, 0.101 & 0.285, 0.252, 0.246 & 13.93 & 128 &  \hyperlink{https://github.com/aehrc/cvt2distilgpt2}{URL}\\
\hline 
\#08  &R2Gen~\cite{Chen2020R2Gen}  &EMNLP20  &Transformer  &Transformer &0.081, 0.246, 0.113, 0.077 & 0.318, 0.200, 0.181 & 110.05 & 83.5 &  \hyperlink{https://github.com/zhjohnchan/R2Gen}{URL}\\
\hline 
\#09  &R2GenCMN~\cite{Chen2021R2GenCMN}  &ACL21  &Transformer  &Transformer &0.087, 0.256, 0.127, 0.102  & 0.329, 0.241, 0.231 & 66.08 & 67.70 &  \hyperlink{https://github.com/zhjohnchan/R2GenCMN}{URL}\\
\hline 
\#10  & Zhu et al.~\cite{Zhu2023LongitudinalChestX-Rays} & MICCAI23 & Transformer & Transformer & 0.074, 0.235, 0.128, 0.078 & 0.217, \underline{0.308}, 0.205 & 10.03 & 85.95 &  \hyperlink{https://github.com/CelestialShine/Longitudinal-Chest-X-Ray}{URL}\\
\hline 
\#11  & CAMANet~\cite{wang2024CAMANet} & IEEE JBH23 & Swin-Former & Transformer &0.083, 0.249, 0.118, 0.090  & 0.328, 0.224, 0.216 & 23.08 & 43.22 &  \hyperlink{https://github.com/Markin-Wang/CAMANet}{URL}\\
\hline 
\#12  & ASGMD~\cite{XUE2024ASGMD} & ESWA24 &\makecell{ResNet-101\\Transformer} & Transformer &0.063, 0.220, 0.094, 0.044 & 0.146, 0.108, 0.055 & 87.37 & 277.41 
&\hyperlink{https://github.com/shangchengLu/ASGMDN}{URL}\\ 
\hline 
\#13  & Token-Mixer~\cite{Yang2024Token-Mixer} & IEEE TMI23 & ResNet-50 & Transformer &0.091, 0.261, 0.135, 0.098  & 0.309, 0.270, \underline{0.288} & 17.54 & 104.34 &  \hyperlink{https://github.com/yangyan22/Token-Mixer}{URL}\\
\hline 
\#14  & PromptMRG~\cite{Jin2024PromptMRG} & AAAI24 & ResNet-101 & Bert &0.095, 0.222, 0.121, 0.044  &0.258, 0.265, 0.281 &108.45  & 219.92  &  \hyperlink{https://github.com/jhb86253817/PromptMRG}{URL}\\
\hline 
\#15  & R2GenGPT~\cite{Wang2023R2GenGPT} & Meta-Rad.23 & Swin-Transformer & Llama2 &0.101, 0.266, 0.145, \underline{0.123} & 0.315, 0.244, 0.260 & 77.8 & 90.9 &  \hyperlink{https://github.com/wang-zhanyu/R2GenGPT}{URL}\\
\hline 
\#16  & WCL~\cite{yan2021WCL} &EMNLP21  &Transformer  &Transformer  &0.084, 0.253, 0.126, 0.103  & 0.335, 0.259, 0.256 & 24.08 & 81.29  &  \hyperlink{https://github.com/zzxslp/WCL}{URL}\\
\hline 
\#17  &R2GenCSR~\cite{wang2024r2gencsr} &arXiv24  & VMamba & Llama2 &0.100, 0.265, 0.146, 0.121 & 0.315, 0.247, 0.259 & 31.2 & 91.7 &  \hyperlink{https://github.com/Event-AHU/Medical_Image_Analysis/tree/main/R2GenCSR}{URL}\\
\hline 
\#18  & VLCI~\cite{chen2024VLCI} & arXiv24 & Transformer & Transformer &0.080, 0.247, 0.114, 0.072  & 0.341, 0.175, 0.163 & 123.71 & 91.46 &  \hyperlink{https://github.com/WissingChen/VLCI}{URL}\\
\hline 
\#19  & Wang et al.~\cite{wang2024pretraininghighdefinitionxray} & arXiv24 & ViT & Llama2 & 0.064, 0.220, 0.110, 0.059 & 0.175, 0.099, 0.078 & 10.82 & 358.80 &  \hyperlink{https://github.com/aehrc/cxrmate}{URL}\\
\hline \toprule [0.5 pt] 
\#20  & MambaXray-VL-B & Ours & MambaXray-VL & Llama2 & \underline{0.105}, \underline{0.267}, \underline{0.149}, 0.117 & 0.333, 0.264, 0.273 & 50.66 & 57.31 &  \hyperlink{https://github.com/Event-AHU/Medical_Image_Analysis}{URL} \\ 
\hline 
\#21  & MambaXray-VL-L  & Ours& MambaXray-VL & Llama2 & \textbf{0.112}, \textbf{0.276}, \textbf{0.157}, \textbf{0.139} & \textbf{0.377}, \textbf{0.319}, \textbf{0.335} & 55.18 & 202.32 &  \hyperlink{https://github.com/Event-AHU/Medical_Image_Analysis}{URL} \\ 
\hline \toprule [0.5 pt]         
\end{tabular}
} 
\label{tab:XraybaseAlgorithms}
\end{table*}

\noindent $\bullet$ \textbf{Fine-tuning Stage.} 
In the downstream fine-tuning stage, we tested the model's performance on three different public datasets. On the IU-Xray~\cite{demner2016iuxray} dataset, we set the maximum training epochs to 30 and the batch size to 20. The visual encoder used was Vim~\cite{zhu2024vim}, loaded with weights from the second stage of pretraining, while the large language model was Qwen-1.5-1.8B~\cite{Bai2023qwen}, with \textit{max\_length} set to 60 and a validation frequency of 1, meaning we validated after each training epoch. On the MIMIC-CXR~\cite{johnson2019mimicCXR} and CheXpert Plus~\cite{chambon2024CheXpertPLUS} datasets, we set the maximum training epochs to 6 and the batch size to 18. The visual encoder remained unchanged, while the large language model used was Llama2-7B~\cite{touvron2023llama2}, with \textit{max\_length} set to 100 and a validation frequency of 0.5, meaning we validated at both the end of each training cycle and after the training was complete. We froze the large language model and trained only the visual encoder and the visual mapper layer, by following the R2GenGPT~\cite{Wang2023R2GenGPT}.

\section{CXPMRG-Bench}  \label{sec::benchmark}
In this paper, we benchmark the newly released CheXpert Plus dataset for the X-ray image based medical report generation. The mainstream MRG algorithms and large language models are listed in the following subsections. For the experimental results, please refer to Table~\ref{tab:XraybaseAlgorithms}, Table~\ref{tab:xrayLLMs}, and Fig.~\ref{fig:firstIMG}.

\subsection{Mainstream MRG Algorithms} 
For the mainstream X-ray image MRG algorithms, as shown in Table~\ref{tab:XraybaseAlgorithms}, we train and test 21 open-sourced algorithms from the year 2020 to the year 2024. These models adopt the 
\textbf{CNN} (ORGan~\cite{hou2023ORGan}, M2KT~\cite{Yang2023M2KT}, ASGMD~\cite{XUE2024ASGMD}, Token-Mixer~\cite{Yang2024Token-Mixer}, PromptMRG~\cite{Jin2024PromptMRG}), 
\textbf{Transformer} (R2GenRL~\cite{qin2022R2GenRL}, XProNet~\cite{wang2022XProNet}, MSAT~\cite{wang2022MSAT}, TIMER~\cite{Wu2023TIMER}, CvT2DistilGPT2~\cite{Nicolson2023CvT2DistilGPT2}, R2Gen~\cite{Chen2020R2Gen}, R2GenCMN~\cite{Chen2021R2GenCMN}, Zhu et al.~\cite{Zhu2023LongitudinalChestX-Rays}, CAMANet~\cite{wang2024CAMANet}, R2GenGPT~\cite{Wang2023R2GenGPT}, WCL~\cite{yan2021WCL}, VLCI~\cite{chen2024VLCI}, Wang et al.~\cite{wang2024pretraininghighdefinitionxray}), and 
\textbf{Mamba} (R2GenCSR~\cite{wang2024r2gencsr}, MambaXray-VL-B, MambaXray-VL-L) as their vision backbone network, and utilize the LSTM, Transformer based model as the decoder network. Note that, the MambaXray-VL-B and MambaXray-VL-L are two models proposed in this paper which will be introduced in the next section.

When reproducing these X-ray based MRG models, we found that some algorithms use \textit{truncated ground truth} for comparison, which we believe may not accurately reflect the true evaluation results. Therefore, we abandoned the truncation mechanism and used the complete ground truth for result evaluation, making the obtained results more accurate and reliable.

\subsection{LLMs for MRG} 
We evaluate a total of 16 open-source LLMs, as shown in Table~\ref{tab:xrayLLMs}, including Vicuna-7B~\cite{zheng2023vicuna}, QWen1.5-7B~\cite{Bai2023qwen}, QWen2-7B-Instruct~\cite{Bai2023qwen}, InternLM-7B~\cite{cai2024internlm2}, Llama2-7B~\cite{touvron2023llama2}, Llama2-13B~\cite{touvron2023llama2}, Llama3-8B~\cite{dubey2024llama3}, Llama3.1-8B~\cite{dubey2024llama3}, GPT2-Medium~\cite{radford2019gpt}, Orca 2-7B~\cite{mitra2023orca}, Orca 2-13B~\cite{mitra2023orca}, DeepSeek-LLM-7B-Chat~\cite{deepseekai2024deepseekllm}, Yi-1.5-6B-Chat~\cite{ai2024yi}, Yi-1.5-9B-Chat~\cite{ai2024yi}. Note that part of the LLMs is selected from \textit{{\textbf{open-llm-leaderboard}}} \footnote{\url{https://huggingface.co/spaces/open-llm-leaderboard/open_llm_leaderboard}} and integrated with R2GenGPT~\cite{Wang2023R2GenGPT} model by replacing the Llama2 language decoder with corresponding LLMs. In our implementation, we keep the visual encoder SwinTransformer unchanged for a fair comparison. In addition, we also test two pre-trained vision-language large models, i.e., InternVL-2~\cite{chen2023internvl} and MiniCPM-V2.5~\cite{yao2024minicpmv}, to check whether a better performance can be obtained, as shown in Table~\ref{tab:xrayLLMs}.

\subsection{Evaluation Results} 
\textbf{[Mainstream MRG Models]} As shown in Table~\ref{tab:XraybaseAlgorithms}, there are five MRG models which achieve a higher B4 metric, i.e., the XProNet~\cite{wang2022XProNet} (0.100), R2GenGPT~\cite{Wang2023R2GenGPT} (0.101), R2GenCSR~\cite{wang2024r2gencsr} (0.100), and our newly proposed MambaXray-VL-B and MambaXray-VL-L which achieves 0.105, and 0.112, respectively. It is intuitive to find that the large language model Llama2 works well for the MRG task. For F1 in the clinical metric, the top-5 models are our newly proposed MambaXray-VL-L (0.335), Token-Mixer~\cite{Yang2024Token-Mixer} (0.288), PromptMRG~\cite{Jin2024PromptMRG} (0.281), ORGan~\cite{hou2023ORGan} (0.277) and our proposed MambaXray-VL-B (0.273). From these results, we can find that our proposed multi-stage pre-training strategy is rather effective in the disease-aware perception of the MRG.

\textbf{[LLM/VLM based MRG Models]} As shown in Table~\ref{tab:xrayLLMs}, we also report the performance of existing widely used LLMs by replacing the Llama2 based on the R2Gen-GPT framework (SwinTransformer is adopted as the vision backbone network). It is easy to find that the Vicuna-V1.5~\cite{zheng2023vicuna} released in the year 2023 achieves the best B4 metric and the InternLM~\cite{cai2024internlm2} performs the best on the F1 clinical metric. 
For the two vision-language models we evaluated, i.e., the InternVL-2 and MiniCPM-V2.5, we can find that their results are not as good as other LLM-based models, although they have similar parameters. These results demonstrate that the vision-language models pre-trained on natural image-pairs may have large gaps with the X-ray medical images. Compared with the mainstream MRG models reported in Table~\ref{tab:XraybaseAlgorithms}, the LLM-based MRG achieves better results than regular language decoders which demonstrates the effectiveness of pre-trained LLMs.

\textbf{[Efficiency \& Parameters]} From the perspective of running efficiency, we test these models on a server with A800 GPUs (80GB). Note that, we set the batch size as large as possible to make full use of the GPU memory. As a result, we can find that MSAT~\cite{wang2022MSAT} and XProNet~\cite{wang2022XProNet} are the first two algorithms that only need 5.72 and 6.3 minutes for the testing subset. R2Gen~\cite{Chen2020R2Gen}, PromptMRG~\cite{Jin2024PromptMRG}, and VLCI~\cite{chen2024VLCI} are relatively slow and need more than 100 minutes on the testing subset of CheXpert Plus dataset. For the LLM-based MRG reported in Table~\ref{tab:xrayLLMs}, we can find that Yi-1.5~\cite{ai2024yi} with 6.1B and 8.8B achieves better efficiency which needs 43.66 and 48.50 minutes for the testing.
From the Fig.~\ref{fig:firstIMG} and Table~\ref{tab:XraybaseAlgorithms}, we can find that the ASGMD~\cite{XUE2024ASGMD}, PromptMRG~\cite{Jin2024PromptMRG}, Wang et al.~\cite{wang2024pretraininghighdefinitionxray}, and our MambaXray-VL-L contains the most parameters (larger than 200M) needed to be tuned in the training phase. However, we can find that our model runs faster than these large models which only need 55.18 minutes. It fully validated the efficiency of our proposed framework for the X-ray image based medical report generation.

\begin{table*}
\centering
\caption{Experimental Results of Medical Report Generation on the CheXpert Plus Dataset using different \textbf{LLMs and VLMs based on R2Gen-GPT}. The symbol $\dagger$ indicates that the model is a VLM. The Param listed in this table denotes the parameters of LLM/VLM.}   
\small 
\resizebox{6.5in}{!}{ 
\begin{tabular}{c|l|l|cccc|ccc|c|c|c} 
\hline  \toprule [0.5 pt] 
\textbf{Index} &\textbf{LLM/VLM} &\textbf{Year}  &\textbf{B4} &\textbf{R} &\textbf{M} &\textbf{C}  &\textbf{P} &\textbf{R} &\textbf{F1} &\textbf{Time (\textit{min})} &\textbf{Param} &\textbf{Code} \\
\hline 
\#01  &Vicuna-V1.5~\cite{zheng2023vicuna} &  2023  & \textbf{0.104} & \textbf{0.272} & \underline{0.160} & \underline{0.202} &  0.334 & 0.258 & \underline{0.276} & 72.00 & 6.7B &   \hyperlink{https://huggingface.co/lmsys/vicuna-7b-v1.5}{URL}\\ 
\hline 
\#02  &Qwen-1.5~\cite{Bai2023qwen} & 2024 &  0.098 & 0.262 & 0.139 & 0.139 & 0.303& 0.233&  0.241& 154.25 & 7.7B &\hyperlink{https://huggingface.co/Qwen/Qwen1.5-7B-Chat}{URL} \\ 
\hline 
\#03  &Qwen-2~\cite{Bai2023qwen} & 2024 & 0.100 & \underline{0.270} & 0.142 & 0.159 & 0.313& \underline{0.269}& 0.261& 103.33 & 7.6B  &\hyperlink{https://huggingface.co/Qwen/Qwen2-7B-Instruct}{URL} \\ 
\hline 
\#04  &InternLM~\cite{cai2024internlm2} & 2024 & 0.063 & 0.207 & 0.136 & 0.104 & 0.307& \textbf{0.274}& \textbf{0.284}& 294.00 & 7.3B &  \hyperlink{https://huggingface.co/internlm/internlm-chat-7b}{URL}\\
\hline 
\#05  &Llama-2~\cite{touvron2023llama2} & 2023 & 0.102 & 0.267 & 0.157 & 0.179 & 0.315& 0.244& 0.260& 77.78 & 6.7B &  \hyperlink{https://huggingface.co/meta-llama/Llama-2-7b-chat-hf}{URL}\\ 
\hline 
\#06  &Llama-2~\cite{touvron2023llama2} & 2023 & 0.101 & 0.269 & \underline{0.160} & \textbf{0.214} & 0.321& 0.254& 0.267& 116.42 & 13.0B &  \hyperlink{https://huggingface.co/meta-llama/Llama-2-13b-chat-hf}{URL}\\ 
\hline 
\#07  &Llama-3~\cite{dubey2024llama3} & 2024 & 0.077 & 0.220 & 0.121 & 0.134 & 0.306& 0.232& 0.222& 130.00 & 8.0B &  \hyperlink{https://huggingface.co/meta-llama/Meta-Llama-3-8B-Instruct}{URL}\\ 
\hline 
\#08  &Llama-3.1~\cite{dubey2024llama3} & 2024 &  0.075 & 0.221 & 0.121 & 0.136 & 0.295& 0.251& 0.242& 110.00 & 8.0B &  \hyperlink{https://huggingface.co/meta-llama/Meta-Llama-3.1-8B-Instruct}{URL} \\ 
\hline 
\#09  &GPT2-Medium~\cite{radford2019gpt} & 2019 & 0.063 & 0.198 & 0.104 & 0.067 & \textbf{0.358}& 0.186& 0.165& 57.33 & 354M &  \hyperlink{https://huggingface.co/openai-community/gpt2-medium}{URL}\\ 
\hline 
\#10  &Orca-2~\cite{mitra2023orca} & 2023 & \underline{0.103} & \underline{0.270} & \textbf{0.161} & 0.199 & 0.330& 0.251& 0.271& 177.33 & 6.7B &  \hyperlink{https://huggingface.co/microsoft/Orca-2-7b}{URL}\\
\hline 
\#11  &Orca-2~\cite{mitra2023orca} & 2023 & 0.100 & 0.266 & 0.159 & 0.187 & 0.317& 0.242& 0.257& 108.66 & 13.0B &  \hyperlink{https://huggingface.co/microsoft/Orca-2-13b}{URL}\\
\hline 
\#12  &Deepseek-LLM~\cite{deepseekai2024deepseekllm} & 2024 & 0.096 & 0.268 & 0.137 & 0.150 & \underline{0.336}& 0.256& 0.253& 201.30 & 6.9B &  \hyperlink{https://huggingface.co/deepseek-ai/deepseek-llm-7b-chat}{URL}\\
\hline 
\#13  &Yi-1.5~\cite{ai2024yi} & 2024 & 0.091 &0.263 & 0.131 & 0.136   & 0.322& 0.229& 0.226& 43.66 & 6.1B &  \hyperlink{https://huggingface.co/01-ai/Yi-1.5-6B-Chat}{URL} \\
\hline 
\#14  &Yi-1.5~\cite{ai2024yi} & 2024 &  0.096 & 0.269 & 0.138 & 0.155 & \underline{0.336}& 0.241&  0.243& 48.50 & 8.8B &  \hyperlink{https://huggingface.co/01-ai/Yi-1.5-9B-Chat}{URL} \\
\hline 
\#15  &InternVL-2\textsuperscript{$\dagger$}~\cite{chen2023internvl} & 2023 &  0.058 & 0.188 & 0.112 & 0.085 & 0.196& 0.127&  0.132& 108.50 & 8.0B &  \hyperlink{https://huggingface.co/OpenGVLab/InternVL2-8B}{URL} \\
\hline
\#16  &MiniCPM-V2.5\textsuperscript{$\dagger$}~\cite{yao2024minicpmv} & 2024 &  0.046 & 0.177 & 0.085 & 0.076 & 0.254& 0.152&  0.122& 51.50 & 8.4B &  \hyperlink{https://huggingface.co/openbmb/MiniCPM-Llama3-V-2_5}{URL} \\
\hline
\toprule [0.5 pt]         
\end{tabular}
} 
\label{tab:xrayLLMs}
\end{table*}

\section{Experiments} \label{sec::experiments}

\subsection{Dataset}  
In the first stage of autoregressive pre-training, we used about 1.27 million medical chest X-ray images proposed in the work~\cite{wang2024pretraininghighdefinitionxray}. In the second stage of image-text contrastive learning pre-training, we used a combination of training data from the \textbf{MIMIC-CXR}~\cite{johnson2019mimicCXR}, \textbf{CheXpert Plus}~\cite{chambon2024CheXpertPLUS}, and \textbf{IU X-ray}~\cite{demner2016iuxray} datasets, totaling 480k image-report pairs. Note that the CheXpert Plus dataset used here consists of images and impressions, not the image and findings combination used in the third stage. We strictly excluded any testing samples used in the third stage, resulting in a total of 210k image-impression pairs. In the third stage, We evaluate the performance of our model on three datasets, including IU X-Ray~\cite{demner2016iuxray}, MIMIC-CXR~\cite{johnson2019mimicCXR}, and CheXpert Plus~\cite{chambon2024CheXpertPLUS} dataset. A brief introduction to these datasets is given below.

\noindent $\bullet$ \textbf{IU X-ray Dataset~}~\cite{demner2016iuxray}~\footnote{\url{https://iuhealth.org/find-medical-services/x-rays}} published in 2016 is one of the most frequently used publicly available medical image datasets for medical report generation. It contains 7,470 images and 3,955 radiology reports, with each report associated with either frontal or both frontal and lateral view images. Each report is divided into four sections: Indication, Comparison, \textit{Findings}, and \textit{Impression}. For a fair comparison, we used the same dataset split protocol as R2GenGPT~\cite{Wang2023R2GenGPT}, dividing the dataset into training, testing, and validation sets with a ratio of 7:1:2.

\noindent $\bullet$ \textbf{MIMIC-CXR Dataset~}~\cite{johnson2019mimicCXR}~\footnote{\url{https://physionet.org/content/mimic-cxr/2.0.0/}} is one of the largest publicly available chest X-ray datasets, containing free-text radiology reports. These records from 2011-2016 include 377,110 radiographic images and 227,835 radiology reports collected from 65,379 patients at the Beth Israel Deaconess Medical Center Emergency Department in Boston, Massachusetts. For fair comparison, we used the same dataset split protocol as R2GenGPT, with 270,790 samples for training the model, and 2,130 and 3,858 samples for validation and testing sets, respectively.

\noindent $\bullet$ \textbf{CheXpert Plus Dataset~}~\cite{chambon2024CheXpertPLUS}~\footnote{\url{https://github.com/Stanford-AIMI/chexpert-plus}} is a new radiology dataset designed to enhance the scale, performance, robustness, and fairness of deep learning models in the field of radiology. This dataset includes 223,228 chest X-rays (in DICOM and PNG formats), 187,711 corresponding radiology reports (de-identified and parsed into 11 sections), de-identified demographic data from 64,725 patients, 14 chest pathology labels, and RadGraph~\cite{Saahil2021RadGraph} annotations. For a fair comparison, we followed the dataset split protocol used in R2GenCSR~\cite{wang2024r2gencsr} which adopted \textit{Findings} as the ground truth and split the training/validation/testing subset based on the ratio 7:1:2. The training subset with 40,463 samples, the validation subset with 5,780 samples, and the testing subset with 11,562 samples.

\subsection{Evaluation Metric} 
For the X-ray medical report generation, we evaluate the model using widely used natural language generation (NLG) metrics, including \textbf{CIDEr}~\cite{vedantam_cider_2015}, \textbf{BLEU}~\cite{papineni_bleu_2002}, \textbf{ROUGE-L}~\cite{lin_rouge_2004}, and \textbf{METEOR}~\cite{banerjee_meteor_2005}. More in detail, 
CIDEr~\cite{vedantam_cider_2015} evaluates text through TF-IDF weighted n-gram matching, placing greater emphasis on the importance of words; 
BLEU~\cite{papineni_bleu_2002} evaluates text quality through n-gram matching; 
ROUGE-L~\cite{lin_rouge_2004} evaluates text using the longest common subsequence; 
METEOR~\cite{banerjee_meteor_2005} improves upon BLEU by considering synonyms and word order.

To measure the accuracy of descriptions for clinical abnormalities, we also report \textbf{Clinical Efficacy (CE) metrics}. CE metrics require the use of the CheXPert~\cite{irvin2019chexpert} toolkit to first extract labels from predictive reports and ground truth, and then to compare the presence status of important clinical observations to capture the diagnostic accuracy of the generated reports. We use \textbf{Precision}, \textbf{Recall}, and \textbf{F1} to evaluate model performance for clinical efficacy metrics.

\begin{table*}[]
\caption{Comparison of our model’s performance on the IU X-ray and MIMIC-CXR datasets. The symbol $\dagger$ indicates that we follow the R2Gen annotation using \textit{Findings} and evaluate with our method, as their report modifies the ground truth to an \textit{Impression} concatenated with \textit{Findings}. The best result is highlighted in bold, and the second-best result is underlined.}
\label{tab:results_iu_mimic}
\resizebox{\linewidth}{!}{
\begin{tabular}{c|l|c|ccccccc}
\hline \toprule [0.5 pt] 
\textbf{Dataset} & \textbf{Methods} & \textbf{Publication} & \textbf{BLEU-1} & \textbf{BLEU-2} & \textbf{BLEU-3} & \textbf{BLEU-4} & \textbf{ROUGE-L} & \textbf{METEOR} & \textbf{CIDEr} \\ \hline
\multirow{13}{*}{\textbf{IU X-Ray}} 
 & R2Gen~\cite{Chen2020R2Gen} & EMNLP 2020 & 0.470 & 0.304 & 0.219 & 0.165 & 0.371 & 0.187 & - \\
 & R2GenCMN~\cite{Chen2021R2GenCMN} & ACL-IJCNLP 2021 & 0.475 & 0.309 & 0.222 & 0.170 & 0.375 & 0.191 & - \\
 & PPKED~\cite{liu2021PPKED} & CVPR 2021 & 0.483 & 0.315 & 0.224 & 0.168 & 0.376 & 0.187 & 0.351 \\
 & AlignTrans~\cite{you2021aligntransformer} & MICCAI 2021 & 0.484 & 0.313 & 0.225 & 0.173 & 0.379 & 0.204 & - \\
 & CMCL~\cite{liu2021CMCL} & ACL 2021 &0.473 & 0.305 & 0.217 & 0.162 & 0.378 & 0.186 & - \\
 & Clinical-BERT~\cite{clinicalBert} & AAAI 2022 & \underline{0.495} & \textbf{0.330} & 0.231 & 0.170 & 0.376 & 0.209 & 0.432 \\
 & METransformer~\cite{Wang2023METransformer} & CVPR 2023 & 0.483 & 0.322 & 0.228 & 0.172 & 0.380 & 0.192 & 0.435 \\
 & DCL~\cite{Li2023DCL} & CVPR 2023 & - & - & - & 0.163 & 0.383 & 0.193 & \textbf{0.586} \\
 & R2GenGPT\textsuperscript{$\dagger$}~\cite{Wang2023R2GenGPT} & Meta Radiology 2023 & 0.465 & 0.299 & 0.214 & 0.161 & 0.376 & \underline{0.219} & \underline{0.542} \\
 % & Token-Mixer~\cite{Yang2024Token-Mixer} & IEEE TMI 2024 & 0.483 & 0.338 & 0.250 & 0.190 & 0.402 & 0.208 & 0.482 \\
 & PromptMRG~\cite{Jin2024PromptMRG} & AAAI 2024 & 0.401 & - & - & 0.098 & 0.160 & \textbf{0.281} & - \\ 
 & BootstrappingLLM~\cite{liu2024bootstrapping} & AAAI 2024 &\textbf{0.499} & \underline{0.323} & \underline{0.238} & \underline{0.184} & \textbf{0.390} & 0.208 & - \\ \cline{2-10} 
 
 & MambaXray-VL-Base & Ours & 0.479 & 0.322 & 0.236 & 0.179 & \underline{0.388} & 0.215 & 0.508 \\
 % & MambaXray-VL-Large & Ours & 0.480 & 0.318 & 0.229 & 0.174 & 0.384 & 0.216 & 0.470 \\
 & MambaXray-VL-Large & Ours & 0.491 & \textbf{0.330} & \textbf{0.241} & \textbf{0.185} & 0.371 & 0.216 & 0.524 \\
 \hline \toprule [0.5 pt] 
\multirow{13}{*}{\textbf{MIMIC-CXR}} 
 & R2Gen~\cite{Chen2020R2Gen} & EMNLP 2020 & 0.353 & 0.218 & 0.145 & 0.103 & 0.277 & 0.142 & - \\
 & R2GenCMN~\cite{Chen2021R2GenCMN} & ACL-IJCNLP 2021 & 0.353 & 0.218 & 0.148 & 0.106 & 0.278 & 0.142 & - \\
 & PPKED~\cite{liu2021PPKED} & CVPR 2021 & 0.360 & 0.224 & 0.149 & 0.106 & 0.284 & 0.149 & 0.237 \\
 & AlignTrans~\cite{you2021aligntransformer} & MICCAI 2021 & 0.378 & 0.235 & 0.156 & 0.112 & 0.283 & 0.158 & - \\
 & CMCL~\cite{liu2021CMCL} & ACL 2021 &0.344 & 0.217 & 0.140 & 0.097 & 0.281 & 0.133 & - \\
 & Clinical-BERT~\cite{clinicalBert} & AAAI 2022 & 0.383 & 0.230 & 0.151 & 0.106 & 0.275 & 0.144 & 0.151 \\
 & METransformer~\cite{Wang2023METransformer} & CVPR 2023 & 0.386 & 0.250 & 0.169 & 0.124 & \textbf{0.291} & 0.152 & \textbf{0.362} \\
 & DCL~\cite{Li2023DCL} & CVPR 2023 & - & - & - & 0.109 & 0.284 & 0.150 & \underline{0.281} \\
 & R2GenGPT\textsuperscript{$\dagger$}~\cite{Wang2023R2GenGPT} & Meta Radiology 2023 & 0.408 & 0.256 & 0.174 & 0.125 & 0.285 & \underline{0.167} & 0.244 \\
 % & Token-Mixer~\cite{Yang2024Token-Mixer} &IEEE TMI 2024 & 0.409 & 0.257 & 0.175 & 0.124 & 0.288 & 0.158 & 0.163 \\
 & PromptMRG~\cite{Jin2024PromptMRG} & AAAI 2024 & 0.398 & - & - & 0.112 & 0.268 & 0.157 & - \\ 
 & BootstrappingLLM~\cite{liu2024bootstrapping} & AAAI 2024 & 0.402 & 0.262 & \underline{0.180} & 0.128 & \textbf{0.291} & \textbf{0.175} & - \\ \cline{2-10} 
 % & R2GenGPT~\cite{R2GenGPT} & Meta Radiology 2023 & 0.411 & 0.267 & 0.186 & 0.134 & 0.297 & 0.160 & 0.269 \\
 
 & MambaXray-VL-Base & Ours & \underline{0.420} & \underline{0.264} & \underline{0.180} & \underline{0.129} & 0.283 & 0.162 & 0.206 \\ 
 & MambaXray-VL-Large & Ours & \textbf{0.422} & \textbf{0.268} & \textbf{0.184} & \textbf{0.133} & \underline{0.289} & \underline{0.167} & 0.241 \\ 
\hline  \toprule [0.5 pt] 
\end{tabular}
}
\end{table*}

\subsection{Comparison with SOTA Algorithms}

\newcommand{\yes}{\textcolor{SeaGreen4}{\ding{51}}}
\newcommand{\no}{\textcolor{DarkRed}{\ding{55}}}

\begin{table*}[h]
\caption{ Component analysis of the key modules in our framework on MIMIC-CXR and CheXpert Plus dataset. The symbol $\dagger$ indicates that we are using the \textit{\textbf{Base}} version of the model, while the others are the \textit{\textbf{Large}} versions. \textbf{Vim-IN1K} indicates the use of weights pre-trained on ImageNet-1K; \textbf{Vim-PTD} indicates the use of weights pre-trained on 1.27 million X-ray images; \textbf{MAE} represents the Masked Auto-encoders pre-training framework; \textbf{ARG} represents the Auto-regressive Generation pre-training framework; \textbf{CTL} represents the contrastive learning loss between images and text; \textbf{SFT} represents supervised fine-tuning. \textbf{B4}, \textbf{R}, \textbf{M}, and \textbf{C} represents BLEU-4, ROUGE-L, METEOR, and CIDEr, respectively.} 
\label{tab:component_iu}
\resizebox{\linewidth}{!}{
\begin{tabular}{l|cccccc|cccc|cccc}
\hline \toprule [0.5 pt] 
\multirow{2}{*}{\textbf{Index}} & \multirow{2}{*}{\textbf{Vim-IN1K}} & \multirow{2}{*}{\textbf{Vim-PTD}} & \multirow{2}{*}{\textbf{MAE}} & \multirow{2}{*}{\textbf{ARG}} & \multirow{2}{*}{\textbf{CTL}} & \multirow{2}{*}{\textbf{SFT}}& \multicolumn{4}{c|}{MIMIC-CXR}   & \multicolumn{4}{c}{CheXpert Plus}  \\ \cline{8-15} 
 & & & & & & & \textbf{B4} & \textbf{R} &  \textbf{M}  & \textbf{C}   & \textbf{B4} & \textbf{R} &  \textbf{M}  & \textbf{C}  \\ \hline
$\#01$ & \no & \no & \no & \no & \no & \yes & 0.125 & 0.285 & \textbf{0.167} & \textbf{0.244}  & 0.101 & 0.266 & 0.145 & 0.123 \\
\hline 
$\#02$ & \yes & \yes & \yes & \no &\no & \yes & 0.104 & 0.260  & 0.141 & 0.154  & 0.094 & 0.257 & 0.140 & 0.104 \\
$\#03$ & \yes & \yes & \no & \yes & \no & \yes & 0.130 & 0.286 &  0.162 & 0.224  & 0.089 & 0.247 & 0.134 & 0.089 \\
\hline 
$\#04$ \textsuperscript{$\dagger$} & \yes & \no & \no & \yes & \no & \yes & 0.108 & 0.264 & 0.144 & 0.170  & 0.090 & 0.249 & 0.132 & 0.103 \\
$\#05$ \textsuperscript{$\dagger$} & \yes & \yes & \no & \yes & \no & \yes & 0.121 & 0.280 & 0.161 & 0.224  & 0.093 & 0.254 & 0.138 & 0.102 \\
$\#06$ \textsuperscript{$\dagger$} & \yes & \yes & \no & \yes & \yes & \yes & 0.129 & 0.283 & 0.162 & 0.206  & 0.105 & 0.267 & 0.149 & 0.117 \\
\hline 
$\#07$ & \yes & \no & \no & \yes & \no & \yes & 0.105 & 0.258 &  0.139 & 0.143  & 0.082 & 0.236 & 0.126 & 0.080 \\
$\#08$ & \yes & \yes & \no & \yes & \no & \yes & 0.130 & 0.286 &  0.162 & 0.224  & 0.089 & 0.247 & 0.134 & 0.089 \\
$\#09$ & \yes & \yes & \no & \yes & \yes & \yes & \textbf{0.133} & \textbf{0.289} & \textbf{0.167} & 0.241  & \textbf{0.112} & \textbf{0.276} & \textbf{0.157} & \textbf{0.139} \\
\hline \toprule [0.5 pt] 
\end{tabular}
}
\end{table*}

\noindent $\bullet$ \textbf{Results on IU X-ray Dataset.~}
As shown in Table~\ref{tab:results_iu_mimic}, it can be seen that both our MambaXray-VL-Base and MambaXray-VL-Large exhibit excellent performance on the IU X-ray dataset. Among them, the MambaXray-VL-Large model is at the SOTA level on BLEU-2 (\textbf{B2}), BLEU-3 (\textbf{B3}), and BLEU-4 (\textbf{B4}) metrics with scores of 0.330, 0.241, and 0.185, respectively. This result indicates the superiority of our method over other report generation methods. However, on some other metrics such as BLEU-1 (\textbf{B1}), ROUGE-L (\textbf{R}), METEOR (\textbf{M}), and CIDEr (\textbf{C}), our method does not achieve optimal performance. This reflects the need to improve the generalization of our method on other datasets.

\begin{table*}
\small 
\centering 
\caption{Comparison of the text encoders used in the second stage on the MIMIC-CXR and CheXpert Plus datasets.} 
\label{tab:BERTvsLlama2}
% \resizebox{\linewidth}{!}{
\begin{tabular}{c|cccc|cccc}
\hline \toprule [0.5 pt] 
\multirow{2}{*}{LLM} & \multicolumn{4}{c|}{MIMIC-CXR}   & \multicolumn{4}{c}{CheXpert Plus}  \\ \cline{2-9} 
 &BLEU-4 &ROUGE-L &METEOR  &CIDEr &BLEU-4 &ROUGE-L &METEOR  &CIDEr  \\ \hline
Baseline  & 0.125 & 0.285 & \textbf{0.167} & \textbf{0.244}  & 0.101 & 0.266 & 0.145 & 0.123 \\
\hline
Llama2~\cite{touvron2023llama2}  & 0.122 & 0.276 & 0.157 & 0.211  & 0.066 & 0.233 & 0.124 & 0.043 \\
\hline
Bio\_ClinicalBERT~\cite{Emily2019Bio_ClinicalBERT}  & \textbf{0.133} & \textbf{0.289}  & \textbf{0.167} & 0.241  & \textbf{0.112} & \textbf{0.276} & \textbf{0.157} & \textbf{0.139} \\
\hline \toprule [0.5 pt] 
\end{tabular}
% }
\end{table*}

\noindent $\bullet$ \textbf{Results on MIMIC-CXR Dataset.~}
As shown in Table \ref{tab:results_iu_mimic}, our method also demonstrates outstanding performance on the MIMIC-CXR dataset, surpasses all other advanced report generation methods, and achieves the most advanced level in several common indicators (e.g., BLEU-1, BLEU-2, BLEU-3, and BLEU-4). Specifically, our method improves the BLEU-4 metric by 6\% compared to R2GenGPT. Encouragingly, we achieved favorable results for two of the three remaining metrics, ROUGE-L and METEOR, with scores of 0.289 for ROUGE-L and 0.167 for METEOR, which again demonstrates the superior performance of our model. In the CIDEr metric, our model achieved a score of 0.241, indicating that MambaXray-VL still has room for improvement.

\noindent $\bullet$ \textbf{Results on CheXpert Plus Dataset.~}
As shown in Table~\ref{tab:XraybaseAlgorithms}, our model MambaXray-VL-Large achieves state-of-the-art performance in all evaluation metric species. These include NLG evaluation metrics and CE evaluation metrics. In detail, for the NLG metrics, our scores on BLEU-4, ROUGE-L, METEOR, and CIDEr are 0.112, 0.276, 0.157, and 0.139, respectively. For the CE metrics, our scores on Precision (\textbf{P}), Recall (\textbf{R}), and F1-score (\textbf{F1}) are 0.377, 0.319, and 0.335, respectively. These experimental results fully demonstrate the superior performance of our model on the CheXpert Plus dataset. In terms of efficiency, our method took 55.18 minutes to complete the testing subset of the CheXpert Plus dataset with a parameter size of 202.32M, showing its effectiveness and efficiency in processing X-ray images.

\subsection{Ablation Study}

\newcommand{\MatchingMamGpt}[1]{\sethlcolor{pink}\hl{#1}}
\newcommand{\MatchingGpt}[1]{\sethlcolor{cyan}\hl{#1}}
\newcommand{\MatchingMam}[1]{\sethlcolor{yellow}\hl{#1}}

\begin{figure*}[h]
\centering
\resizebox{\textwidth}{!}{
\begin{tabular}{lp{8cm}p{8cm}p{8cm}}
\hline \toprule [0.5 pt] 
\multicolumn{1}{c}{\textbf{Image}} & \multicolumn{1}{c}{\textbf{Ground Truth}} & \multicolumn{1}{c}{\textbf{Ours}} & \multicolumn{1}{c}{\textbf{R2GenGPT}} \\ 
\hline \toprule [0.5 pt]

\begin{minipage}[b]{0.4\columnwidth}\vspace{7pt}\centering\raisebox{-.9\height}{\includegraphics[width=\linewidth]{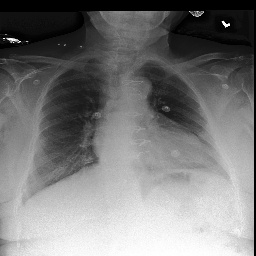}}\end{minipage}& Pa \MatchingMam{and lateral views of the chest provided.} \MatchingMam{Midline sternotomy wires and mediastinal clips are again noted.} The previously noted port-a-cath has been removed. The lungs are clear bilaterally without \MatchingMamGpt{focal consolidation effusion or pneumothorax.} \MatchingMam{Cardiomediastinal} \MatchingMamGpt{silhouette is stable}. \MatchingMam{Bony structures are intact.} \MatchingMam{No free air below the right hemidiaphragm is seen.}
 & Ap upright \MatchingMam{and lateral views of the chest provided.} \MatchingMam{Midline sternotomy wires and mediastinal clips are again noted.} There is no \MatchingMam{focal consolidation} large \MatchingMam{effusion or pneumothorax.} The \MatchingMam{cardiomediastinal silhouette is stable.} \MatchingMam{Bony structures are intact.} \MatchingMam{No free air below the right hemidiaphragm is seen.} 
 & Frontal and lateral views of the chest were obtained. The patient is status post median sternotomy and cabg. The cardiac and mediastinal \MatchingGpt{silhouettes are stable.} There is no \MatchingGpt{focal consolidation} pleural \MatchingGpt{effusion or pneumothorax.} Mild pulmonary vascular congestion is noted. Degenerative changes are seen in the thoracic spine. \\ 

\begin{minipage}[b]{0.4\columnwidth}\vspace{7pt}\centering\raisebox{-.9\height}{\includegraphics[width=\linewidth]{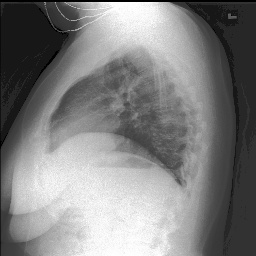}}\end{minipage} & \MatchingMamGpt{Pa and lateral views of the chest provided. There is no focal consolidation effusion or pneumothorax. The cardiomediastinal silhouette is normal. Imaged osseous structures are intact. No free air below the right hemidiaphragm is seen.} Elevation of the right hemidiaphragm is unchanged from chest radiograph.
 & \MatchingMam{Pa and lateral views of the chest provided. There is no focal consolidation effusion or pneumothorax. The cardiomediastinal silhouette is normal. Imaged osseous structures are intact. No free air below the right hemidiaphragm is seen.} Clips are noted in the right upper quadrant of the abdomen. 
 & \MatchingGpt{Pa and lateral views of the chest provided. There is no focal consolidation effusion or pneumothorax. The cardiomediastinal silhouette is normal. Imaged osseous structures are intact. No free air below the right hemidiaphragm is seen.} Surgical clips in the right upper quadrant suggest prior cholecystectomy .\\ 

\begin{minipage}[b]{0.4\columnwidth}\vspace{7pt}\centering\raisebox{-.9\height}{\includegraphics[width=\linewidth]{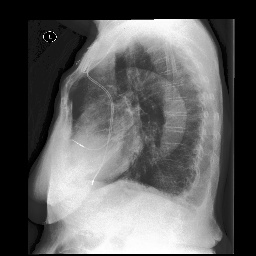}}\end{minipage} &  \MatchingMam{Frontal and lateral views of the chest were obtained. Dual-lead left-sided pacemaker is again seen with leads extending to the expected positions of the right atrium and right ventricle.} The lungs are hyperinflated with \MatchingMamGpt{flattening of the diaphragms} suggesting \MatchingMamGpt{chronic obstructive pulmonary disease.} \MatchingMamGpt{No pleural effusion or pneumothorax} is seen. Slight increased opacity at the right lung base best seen on the fron.
 & \MatchingMam{Frontal and lateral views of the chest were obtained. Dual-lead left-sided pacemaker is again seen with leads extending to the expected positions of the right atrium and right ventricle.} There is \MatchingMam{no} evidence of \MatchingMam{pneumothorax or pleural effusion}. \MatchingMam{The lungs are hyperinflated with flattening of the diaphragms} consistent with \MatchingMam{chronic obstructive pulmonary disease.} Cardiomediastinal silhouette is stable. Bony structures are intact. 
 & Frontal and lateral chest radiographs demonstrate hyperexpanded lungs with \MatchingGpt{flattening of the diaphragms} consistent with \MatchingGpt{chronic obstructive pulmonary disease.} There is \MatchingGpt{no} focal consolidation \MatchingGpt{pleural effusion or pneumothorax.} The cardiac mediastinal and hilar contours are unremarkable. A left-sided pacemaker device is noted with leads terminating in the right atrium and right ventricle. \\ 

\begin{minipage}[b]{0.4\columnwidth}\vspace{7pt}\centering\raisebox{-.9\height}{\includegraphics[width=\linewidth]{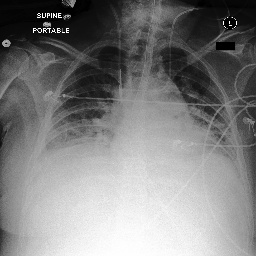}}\end{minipage}\vspace{7pt}&  \MatchingMam{As compared to the previous radiograph there is no relevant change.} \MatchingMamGpt{The monitoring and support devices} \MatchingMam{are} constant. Low lung volumes \MatchingMam{borderline size of the cardiac silhouette.} Mild pulmonary edema. Moderate retrocardiac atelectasis. \MatchingGpt{No evidence of pneumonia}.
 & \MatchingMam{As compared to the previous radiograph there is no relevant change.} \MatchingMam{The monitoring and support devices are} in unchanged position. Low lung volumes with minimal atelectasis at both lung bases. No larger pleural effusions or pneumothorax. \MatchingMam{Borderline size of the cardiac silhouette}. No pulmonary edema. No other parenchymal abnormalities. 
 & In comparison with the study of \MatchingGpt{the monitoring and support devices} remain in place. Continued enlargement of the cardiac silhouette with pulmonary vascular congestion and bilateral pleural effusions with compressive atelectasis at the bases. \MatchingGpt{No evidence of} acute focal \MatchingGpt{pneumonia} or pneumothorax. Central catheters remain in place. \\ 
\hline \toprule [0.5 pt] 
\end{tabular}
}
% \caption{X-ray image and its corresponding ground truth, along with the output of our model generation report on the MIMIC-CXR dataset. The match sentence in the reports are highlighted using different colors.}
\caption{X-ray images and their corresponding ground-truths, along with the output of our model and R2GenGPT model generation reports on the MIMIC-CXR dataset. Matching sentences in our report are highlighted in yellow, R2GenGPT matching sentences are highlighted in cyan, and sentences matching by both models are highlighted in pink.}
\label{fig:visual_report}
\end{figure*}

\noindent $\bullet$ \textbf{Effectiveness of Autoregressive Generation for Pre-training on X-ray Image?}
% compare with MAE 
As shown in Table~\ref{tab:component_iu}, we first compare the autoregressive generation (ARG) pre-training with the Masked Auto-Encoder (MAE) pre-training. From the $\#02$ and $\#03$ rows, it can be seen that the results achieve 0.130/0.089 on the BLEU-4 metric of the MIMIC-CXR and CheXpert Plus datasets, respectively. Note that the ARG pre-training method outperforms the MAE on all metrics, with a +45\% (i.e., (0.224-0.154)/0.154) improvement on CIDEr compared to MAE. The ARG-based pre-training achieves similar performance compared with MAE-based pre-training on the CheXpert Plus dataset.

\noindent $\bullet$ \textbf{Effectiveness of Xray-Report Contrastive Learning.}  
In addition, we further explored the impact of contrastive learning (CTL) on the final performance. The experimental results in the $\#05$ and $\#06$ rows of Table~\ref{tab:component_iu} demonstrate its effectiveness. After introducing the CTL loss, we find that the results on the MIMIC-CXR and CheXpert Plus datasets have all received improvement. More in detail, it improves the ROUGE-L metric by over +5\% on the CheXpert Plus dataset. These experiments demonstrate the positive effect of the CTL loss we used in the pre-training stage.

\noindent $\bullet$ \textbf{Comparison between ViT and Mamba using Autoregressive Generation.} 
As shown in the $\#01$ and $\#09$ rows of Table~\ref{tab:component_iu}, the $\#01$ row uses a visual coder based on the Transformer architecture, while the last row uses a visual coder with auto-regressive pre-training of the Mamba architecture. It can be clearly observed that the encoder based on the Mamba architecture achieves better performance in the vast majority of metrics, both on the MIMIC-CXR and CheXpert Plus datasets, especially on BLEU-4 for the MIMIC-CXR data, where the Mamba architecture improves by +6\% compared to the Transformer architecture. However, on the MIMIC-CXR dataset, the metric CIDEr does not score significantly better than the Transformer architecture. Overall, this series of experiments is sufficient to demonstrate the effectiveness of the auto-regressive pre-trained visual coder based on the Mamba architecture.

\noindent $\bullet$ \textbf{Clinical-BERT vs Llama2 in Xray-Report Contrastive Learning.}  
In this work, we test two models for contrastive learning in the second stage, i.e., the Bio\_ClinicalBERT~\cite{Emily2019Bio_ClinicalBERT} and Llama2~\cite{touvron2023llama2}. As shown in Table~\ref{tab:BERTvsLlama2}, the experimental results on both MIMIC-CXR and CheXpert Plus datasets all demonstrate that the Bio\_ClinicalBERT~\cite{Emily2019Bio_ClinicalBERT} achieves a better performance for the X-ray report generation. We think this may be caused by the fact that the Bio\_ClinicalBERT~\cite{Emily2019Bio_ClinicalBERT} is an LLM pre-trained using medical data, while the Llama2~\cite{touvron2023llama2} is pre-trained using common text data and sensitive to parameter tuning. This experiment inspired us to consider pre-training large language models using medical data in future works.

\noindent $\bullet$ \textbf{Analysis on Different Configurations of Mamba Vision Encoder.}  
% MambaXray-VL-Base vs MambaXray-VL-Large 
Intuitively, the large version of the Mamba model has better generalization and robustness compared to the base version, as it has deeper network layers or higher feature dimensions. As shown in Table~\ref{tab:component_iu}, we can see that the results in lines $\#7$, $\#8$, and $\#9$ (Vim-large) are significantly better than lines $\#4$, $\#5$, and $\#6$ (Vim-base). Meanwhile, our Vim-large achieved optimal performance in experiments after equipping all modules. Thus, it is obvious that the larger version of Vim has a more stable performance on both MIMIC-CXR and CheXpert Plus datasets.

\noindent $\bullet$ \textbf{Does VLMs Pre-trained using Natural Image-Text Samples Ready for the X-ray Report Generation?} In this paper, we also conduct supervised fine-tuning on the CheXpert Plus dataset using Vision-Language Models (VLMs), including InternVL-2~\cite{chen2023internvl} and MiniCPM V2.5~\cite{yao2024minicpmv}. We replace the vision and language backbone network of R2Gen-GPT using the VLMs to adapt them for the X-ray image based report generation task. As illustrated in Table \ref{tab:xrayLLMs}, we can find that the performance of the two models is not as good as the compared models. These experiments demonstrate a large gap between pre-training on the natural and X-ray images. In our future works, we consider further adapting the pre-trained VLMs using natural images to the X-ray image domain to achieve a better performance.

\subsection{Visualization} 
As shown in Fig.~\ref{fig:visual_report}, we give some examples to illustrate the effectiveness of our proposed MambaXray-VL model for the X-ray image based report generation. 
For specific X-ray images, we compared ground truth with the report generated by the MambaXray-VL model and the report generated by the R2GenGPT model. The X-ray images we chose contain both front and side views, normal images, and images containing lesion areas, enabling a more comprehensive and rational visualization. For a more intuitive visualization, we have highlighted the parts that match the ground truth. The yellow highlighted area is the part of the report generated by our model that matches the ground truth, and the blue highlighted area is the part of the report generated by the R2GenGPT model that matches the ground truth. The pink highlighted area is the portion of the report generated by both our model and the R2GenGPT model that matches the ground truth. It is clear that the report generated by our model is closer to the real report than the report generated by the R2GenGPT model, which indicates that our model is effective.

\subsection{Limitation Analysis} 
This paper provides a comprehensive benchmark for the X-ray image based medical report generation, which covers the mainstream MRG models and LLMs. The LLMs evaluated in this work focus on 7B and 13B which is hardware friendly, and the LLMs with more parameters are not discussed due to the limited computational resources. On the other hand, there are still many Vision-Language Models (VLMs) developed for natural images that are not benchmarked, due to the limited performance of the X-ray image-based medical report generation.

\section{Conclusion and Future Works}  \label{sec::conclusion} 
In this work, we propose to benchmark the CheXpert Plus dataset by re-training the mainstream X-ray report generation models and large language models. This benchmark will help identify which large models and algorithms are leading in this domain, significantly promoting academic progress and technological development. In addition, we also propose a new Mamba-based vision-language large model for the X-ray image based medical report generation. It involves three pre-training stages which make full use of auto-regressive generation loss, Xray-report contrastive learning, and supervised fine-tuning. We validate the effectiveness of our proposed pre-trained large model on IU X-ray, MIMIC-CXR, and CheXpert Plus datasets. From the newly built benchmark, we can find that the current large language models still perform poorly on the report generation task.

In our future works, we will consider introducing structured knowledge graphs into the large language model to guide the report generation. In addition, fine-grained X-ray image patch mining guided by the medical report may be another idea worthy of study. We leave them as the future works.

{
    \small
    \bibliographystyle{ieeenat_fullname}
    \bibliography{reference}
}

\end{document}